\documentclass[conference]{IEEEtran}
\IEEEoverridecommandlockouts
\usepackage[backend=biber, style=numeric, sorting=none]{biblatex} 
\usepackage{amsmath,amssymb,amsfonts}
\usepackage{algorithmic}
\usepackage{array, multirow,graphicx}
\usepackage{textcomp}
\usepackage{lipsum}
\usepackage{xcolor}
\def\BibTeX{{\rm B\kern-.05em{\sc i\kern-.025em b}\kern-.08em
    T\kern-.1667em\lower.7ex\hbox{E}\kern-.125emX}}
\begin{document}

\title{A Multimodal Intermediate Fusion Network with Manifold Learning for Stress Detection}
\author{\IEEEauthorblockN{1\textsuperscript{st} Morteza Bodaghi}
\IEEEauthorblockA{\textit{Dept. MCHE. Engineering} \\
\textit{University of Louisiana}\\
Lafayette, USA \\
Bodaghi.Morteza@gmail.com}
\and
\IEEEauthorblockN{2\textsuperscript{nd} Majid Hosseini}
\IEEEauthorblockA{\textit{Dept. C. Science } \\
\textit{University of Louisiana}\\
Lafayette, USA \\
Mjhoseiny@gmail.com}
\and
\IEEEauthorblockN{3\textsuperscript{rd} Raju Gottumukkala}
\IEEEauthorblockA{\textit{Dept. MCHE. Engineering} \\
\textit{University of Louisiana}\\
Lafayette, USA \\
Raju.Gottumukkala@louisiana.edu}
}

\maketitle

\begin{abstract}
Multimodal deep learning methods capture synergistic features from multiple modalities and have the potential to improve accuracy for stress detection compared to unimodal methods. However, this accuracy gain typically comes from high computational cost due to the high-dimensional feature spaces, especially for intermediate fusion. Dimensionality reduction is one way to optimize multimodal learning by simplifying data and making the features more amenable to processing and analysis, thereby reducing computational complexity. This paper introduces an intermediate multimodal fusion network with manifold learning-based dimensionality reduction. The multimodal network generates independent representations from biometric signals and facial landmarks through 1D-CNN and 2D-CNN. Finally, these features are fused and fed to another 1D-CNN layer, followed by a fully connected dense layer. We compared various dimensionality reduction techniques for different variations of unimodal and multimodal networks. We observe that the intermediate-level fusion with the Multi-Dimensional Scaling (MDS) manifold method showed promising results with an accuracy of 96.00\% in a Leave-One-Subject-Out Cross-Validation (LOSO-CV) paradigm over other dimensional reduction methods. MDS had the highest computational cost among manifold learning methods. However, while outperforming other networks, it managed to reduce the computational cost of the proposed networks by 25\% when compared to six well-known conventional feature selection methods used in the preprocessing step.
\end{abstract}

\begin{IEEEkeywords}

Psychological Stress Detection, Early Fusion, Intermediate Level Fusion, Manifold Learning, Dimension Reduction, Multimodal Learning, Biometric Signals, Facial Landmarks, video

\end{IEEEkeywords}

\section{Introduction}
\label{introduction}

 Predicting stress in the real world is a very challenging problem because of the high variability of how stress manifests across individuals and the influence of environmental context. \cite{li2020stress}.
 Multimodal learning integrates signals from multiple modalities and captures complementary information across sources and their cross-modal interaction. Prior work in multi-modal learning from biometric signals, voice, and facial expression has been shown to improve 
 prediction accuracy \cite{li2020stress}. 
 However, as pointed out by prior research \cite{hosseini2023multimodal}, multimodal learning models come with high computational costs of data processing, training, and validation. Training the model requires synchronization across multiple modalities, and optimizing the model to capture the correlation between different modalities gets quite expensive\cite{hosseini2022multimodal}. More research is needed to optimize the performance trade-offs \cite{kapsouras2020deep,zhao2023stress,zhang2019multimodal}. Various strategies exist to optimize the performance trade-offs between achieving good accuracy and computational needs, such as improving network architecture \cite{ugalde2015computational}, model compression \cite{zhu2023survey},  dimensionality reduction\cite{wang2012folded}, regularization\cite{murugan2017regularization} and hardware optimization \cite{al2021review}. In this paper, we study the problem of dimensionality reduction using manifold learning.

Dimensional reduction benefits multimodal learning in various ways. First, unnecessary features are eliminated, and a cleaner set of features is retained. Second, the data size is reduced, reducing the storage and processing requirements\cite{nguyen2019ten} that decreases memory and computational cost. Finally, a reduced feature set minimizes the number of dimensions and redundancy, thereby minimizing overfitting and improving model performance \cite{al2019redundancy}. Existing methods, such as PCA, offer good dimensionality reduction, but these models work well for transformations where the combination of features is predominantly linear. Emotion and stress-related features involve complex geometric transformations that are inherently nonlinear \cite{huang2012nonlinear, akhloufi2009multispectral}. Manifold learning assumes that the data lies on a two-dimensional manifold embedded in a higher-dimensional space, and the structure can be preserved by geometric space, so Manifold learning can potentially capture low-dimensional structures within high-dimensional data where the structure is inherently geometric.  Manifold learning has been applied in image recognition \cite{hoffmann2020learning}, multiple sensor translation \cite{dutt2022shared}, neuroscience and brain network analysis \cite{zhu2019multimodal,song2020multi}. In this paper, we introduce a low-cost intermediate fusion model for stress detection that implements manifold dimensional reduction and convolutional neural network. We investigate different variations of multimodal learning techniques with various manifold learning techniques and PCA to study the accuracy vs computational cost trade-offs. The models are investigated using the LOSO-CV method. We also compared the performance\/cost trade-offs of using manifold learning in unimodal, early, and intermediate-level fusion neural networks.

We discuss closely related work in this area in Section 2; the details of the proposed intermediate fusion network are discussed in Section 3; manifold learning methods are explained in Section 4, and then we discuss results and analysis in Section 5. Section 6 compares the computational cost of different networks with and without using various dimensionality reduction techniques. Finally, we discuss the limitations and conclusion in sections 7 and 8.

\section{Related Work}
Multimodal stress detection is a valuable approach, as it leverages multiple sensory data sources to uncover intriguing patterns and correlations across different modalities simultaneously \cite{walambe2023employing}. In this regard, researchers have explored three distinct classes of multimodal learning models: early fusion, late fusion, and intermediate-level fusion.

Early fusion has been extensively examined in existing literature. For instance, Wu, Yujin, et al. \cite{wu2022fusion} proposed an innovative geometric multimodal LSTM-based stress detection framework, incorporating representations of physiological and behavioral signals derived from covariance and cross-covariance. Similarly, Seo, Wonju, et al. \cite{seo2022deep} introduced a deep learning feature-level (early-fusion) fusion network that combined ECG, respiration, and facial features, achieving a binary stress detection accuracy of 73.3\%. Additionally, \cite{zhang2022multimodal} employed ECG signals, peripheral physiological signals, and eye movement for emotion detection, reducing data dimensions through the ISO algorithm at the feature level and subsequently feeding it into a Convolutional Neural Network (CNN), achieving an accuracy of 90.05

Moving to intermediate-level fusion, \cite{radhika2021deep} attained an accuracy of 75.5\% using a multimodal fusion model based on CNNs for subject-independent stress detection. This model incorporated physiological signals, such as Electrocardiogram (ECG) and Electrodermal Activity (EDA). Furthermore, \cite{sreevidya2022elder} successfully fused raw video and audio signals to create an emotion detection system tailored for the elderly population, showing significant improvements in detecting happiness and sadness.

Other approaches integrated various modalities for stress detection. For instance, \cite{zhang2022real} integrated ECG, voice, and facial expression data, achieving an accuracy of 85.1\% in stress detection. Meanwhile, \cite{zali2023deep} proposed subject-independent emotion recognition using EEG signals, employing a Semi-Supervised Dimension Reduction algorithm (SSDR) to reduce deep feature dimensions and improve classification accuracy. Notably, fusion after profound feature reduction yielded the highest accuracy. Lastly, \cite{she2023cross} combined EEG and speech features for emotion recognition, employing a weighted fusion approach and achieving an average accuracy of 86.59\%.

These studies underscore the significance of employing multiple modalities for stress detection, as they enable the integration of complementary information. However, there remains a need for improved, more accurate, and faster stress detection methods suitable for real-world scenarios, facilitating timely stress monitoring and management.

This paper focuses on exploring the computational cost and performance trade-offs of various Manifold-based dimensional reduction methods within an intermediate-level fusion network. We compare these methods to unimodal and early fusion networks for stress detection. Among these dimensionality reduction techniques, our experiments reveal that Multidimensional Scaling (MDS) achieves a remarkable accuracy of 96\% when utilized in an intermediate-level fusion network, particularly in a Leave-One-Subject-Out Cross-Validation (LOSO-CV) framework.

\section{Multimodal Intermediate Level Fusion Network}

\begin{figure}[h!]
    \centering
    \includegraphics[width = \linewidth]{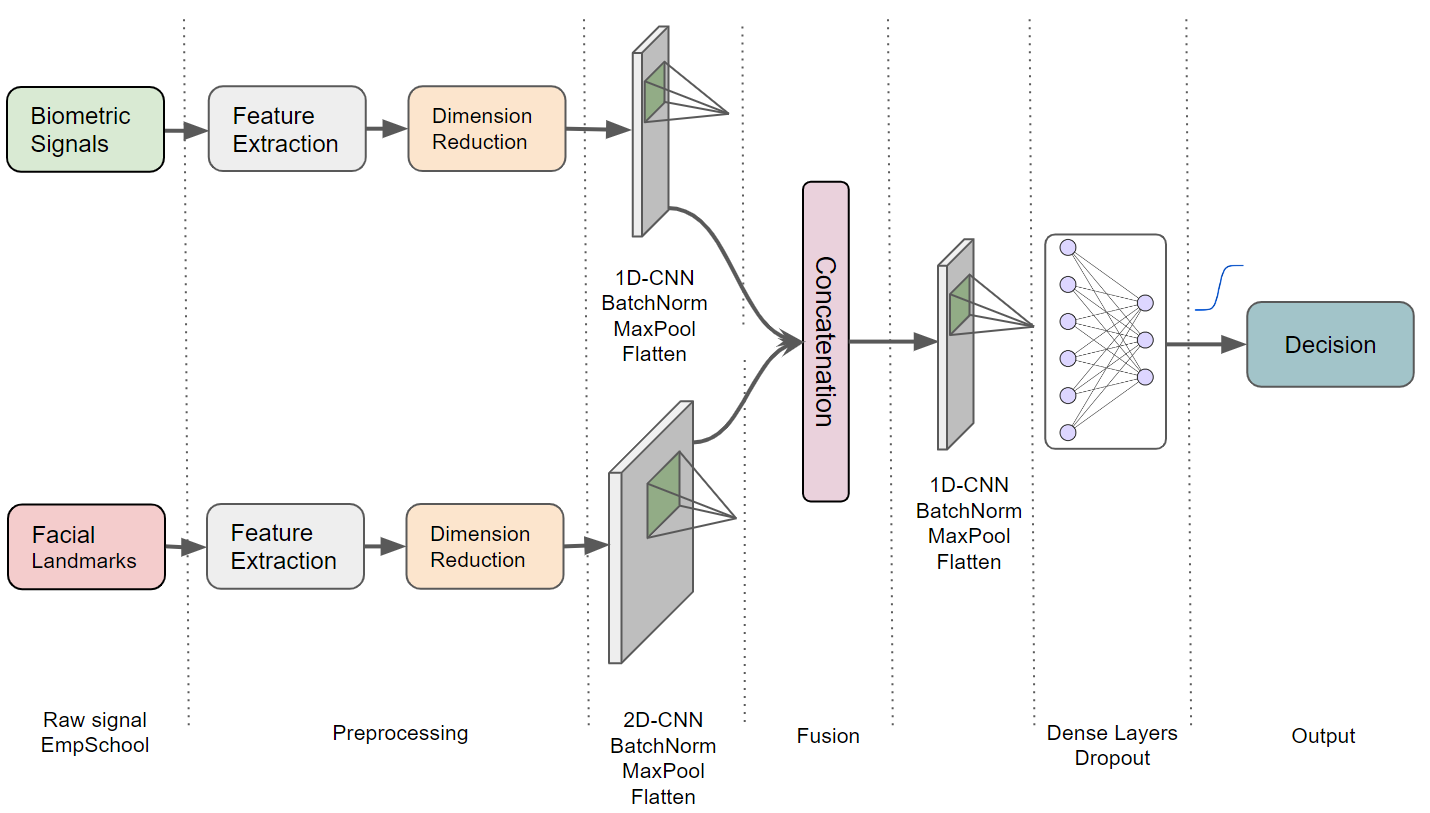}
    \caption{Intermediate Level Fusion Network}
    \label{fig:net}
\end{figure}

Figure \ref{fig:net} illustrates our proposed intermediate fusion network. This model employs manifold dimensional reduction before the fusion of the representations of each modality separately. 

In this network architecture, the process begins with the preprocessing of biometric signals and facial landmarks. The biometric signals are input into a $3\times1$ CNN layer, while in parallel, a $3\times3$ CNN layer is responsible for generating representations of facial landmarks. Subsequently, both sets of representations go through $(2\times1)$ and $(2\times2)$ max-pooling layers, followed by flattening layers.

The representations derived from biometric signals and facial landmarks are then combined through concatenation before being fed into a series of fully connected Deep Neural Network (DNN) layers. The DNN layers consist of the following components: A fully connected layer with 16 neurons employing a ReLU activation function, followed by a 0.2 dropout, and a layer with eight neurons utilizing ReLU activation, L2 regularization with a coefficient of 0.01, followed by a 0.2 dropout. The final output layer comprises three neurons and employs a softmax activation function for the purpose of classifying stress into three distinct levels.

The source code and network architectures related to this work are made publicly available through the CPHS lab at the University of Louisiana at Lafayette \cite{CPHSLAB}.

\section{Manifold learning based Dimensionality Reduction}
Manifold learning is a technique that uncovers the underlying structure or geometry within data and then transforms it into a space with fewer dimensions. This approach allows us to gain insights into high-dimensional data either by projecting it onto a lower-dimensional space or by learning the mapping itself. By identifying a non-linear mapping that connects the high-dimensional space to a lower-dimensional manifold, manifold learning preserves the inherent relationships between the data points while simultaneously simplifying computational complexity \cite{ma2011manifold}. We employed six different dimension reduction methods, including Locally Linear Embedding (LLE)\cite{roweis2000nonlinear}, Spectral Embedding (SE)\cite{belkin2003laplacian}, Multidimensional Scaling (MDS)\cite{borg2005modern}, Isometric Mapping (ISO)\cite{tenenbaum2000global}, t-Distributed Stochastic Neighbor Embedding (t-SNE)\cite{van2008visualizing}, and Principal Component Analysis (PCA)\cite{jolliffe2016principal}. 
The following sections explain various manifold learning techniques. 

\subsection{Locally Linear Embedding (LLE)}
The LLE method, introduced by \cite{roweis2000nonlinear}, is an unsupervised dimensionality reduction technique that attempts to retain the essential geometric characteristics of the original nonlinear feature structure. The method aims to find a lower-dimensional representation of the data while ensuring that the distances between the k-nearest neighbors are maintained. It achieves this by modeling the data as a linear combination of its neighboring points and then determining the weights that minimize the dissimilarity between the original data and its reconstructed version.

\begin{figure}[h!]
    \centering
    \includegraphics[width = \linewidth]{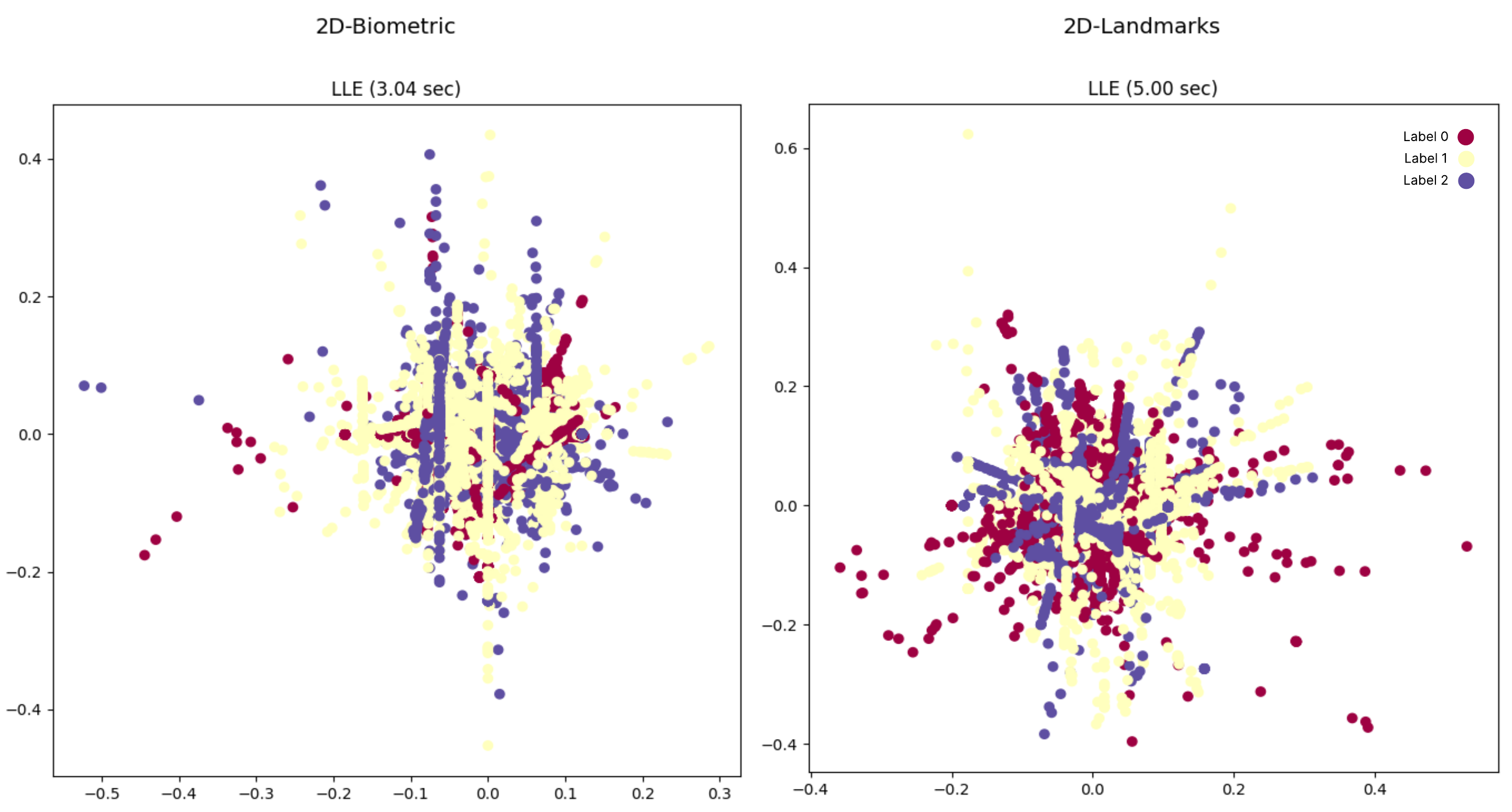}
    \caption{Biometric and Landmarks data visualization using LLE}
    \label{fig:fig1}
\end{figure}

To illustrate the performance of LLE, we provide a 2D visualization of our data in Figure \ref{fig:fig1}, along with the time it took to reduce their dimension using LLE. The plots clearly demonstrate that LLE encountered challenges in effectively clustering the data while preserving the local structure. For this visualization, we employed the following parameters: Number of neighbors = 10, Number of Components = 2, method = 'standard', and random state = 42.

\subsection{Spectral Embedding (SE)}
\begin{figure}[h!]
    \centering
    \includegraphics[width = \linewidth]{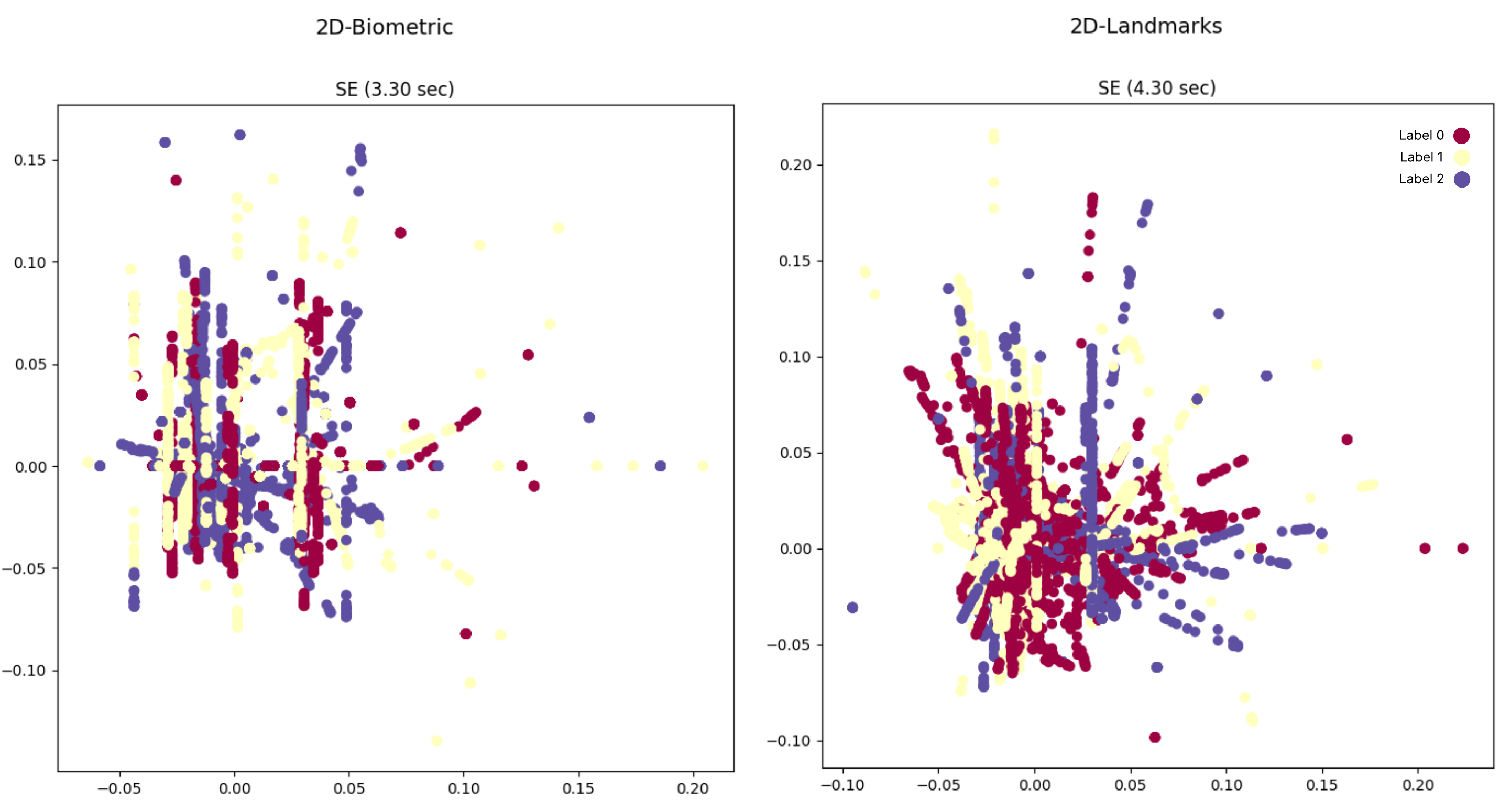}
    \caption{Biometric and Landmarks data visualization using SE}
    \label{fig:fig3}
\end{figure}

SE is another manifold nonlinear dimension reduction technique introduced in 2002 by \cite{belkin2003laplacian}. SE simplifies complex data relationships by analyzing the connections between points using Laplacian Eigenmaps and creating a smaller data version. This helps to understand complex relationships that do not follow a straight line. To visualize our data in 2D using SE, we used the following parameters: Number of neighbors = 5, Number of Components = 2, and random state = 42. Figure \ref{fig:fig3} shows the 2D visualization of our data using SE. As shown in the plots, SE exhibited nonsensical clustering in the data and yielded the worst performance due to its limitation in capturing only the global aspect of the data, making it unsuitable for our dataset.

\subsection{Multi-Dimensional Scaling (MDS)}
\begin{figure}[h!]
    \centering
    \includegraphics[width = \linewidth]{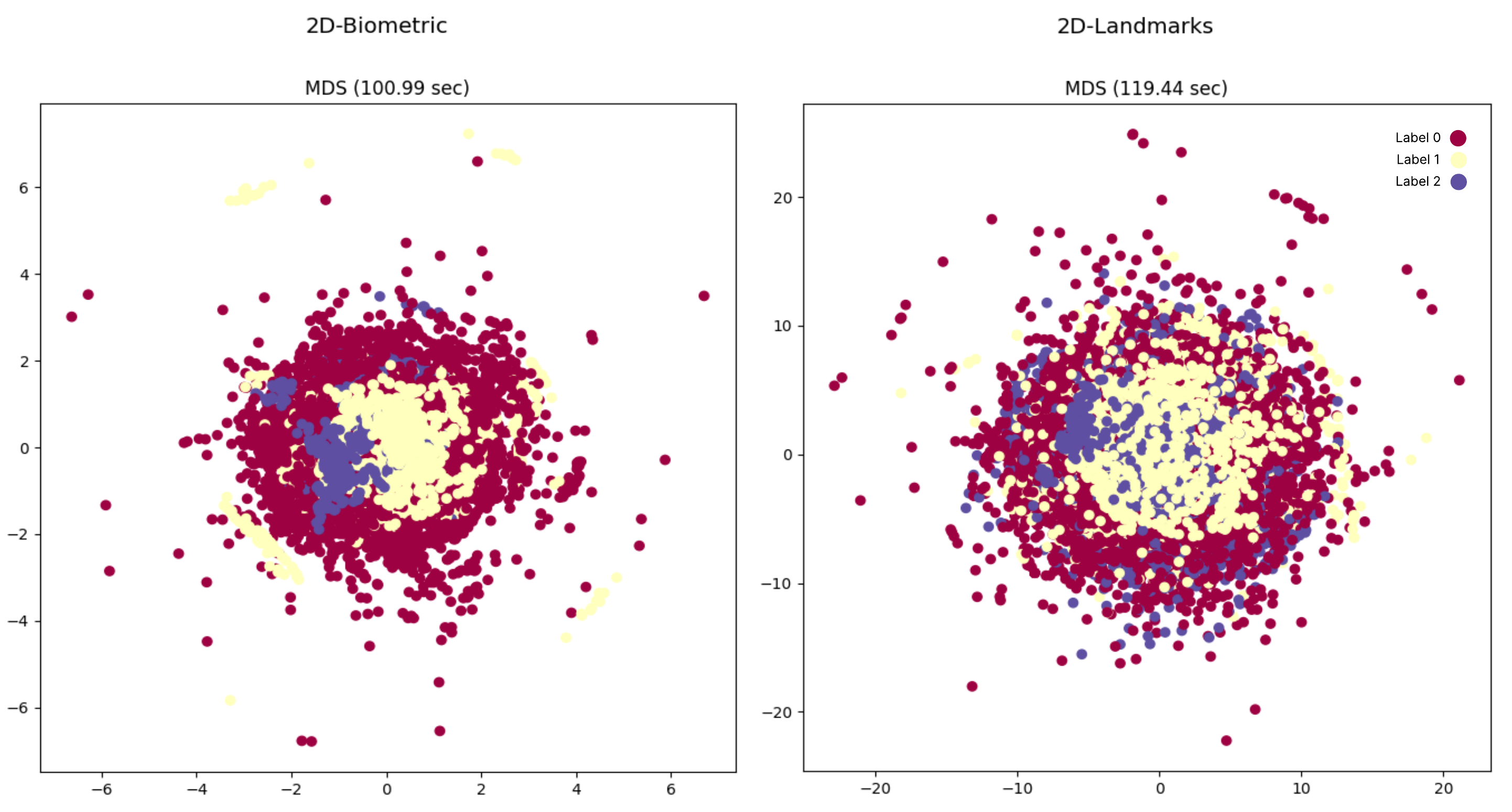}
    \caption{Biometric and Landmarks data visualization using MDS}
    \label{fig:fig6}
\end{figure}

MDS measures distances or similarities between data points in high-dimensional data and generates a new map that accurately retains these relationships. This method captures a strong local structure of the data based on pairwise distances between the data points \cite{borg2005modern}. The visualization of our data using 2D MDS is illustrated in Figure \ref{fig:fig6}. We used the following parameters for visualization: Number of Components = 2 and random state = 42. Our study shows that MDS plots effectively cluster the data, especially for biometric signals, making it the best-performing technique in this study when using intermediate-level fusion. However, the downside of MDS is its high computational cost, which needs to be considered. Despite this, MDS is a powerful method with notable strengths.

\subsection{Isometric Mapping (ISO)}
\begin{figure}[h!]
    \centering
    \includegraphics[width = \linewidth]{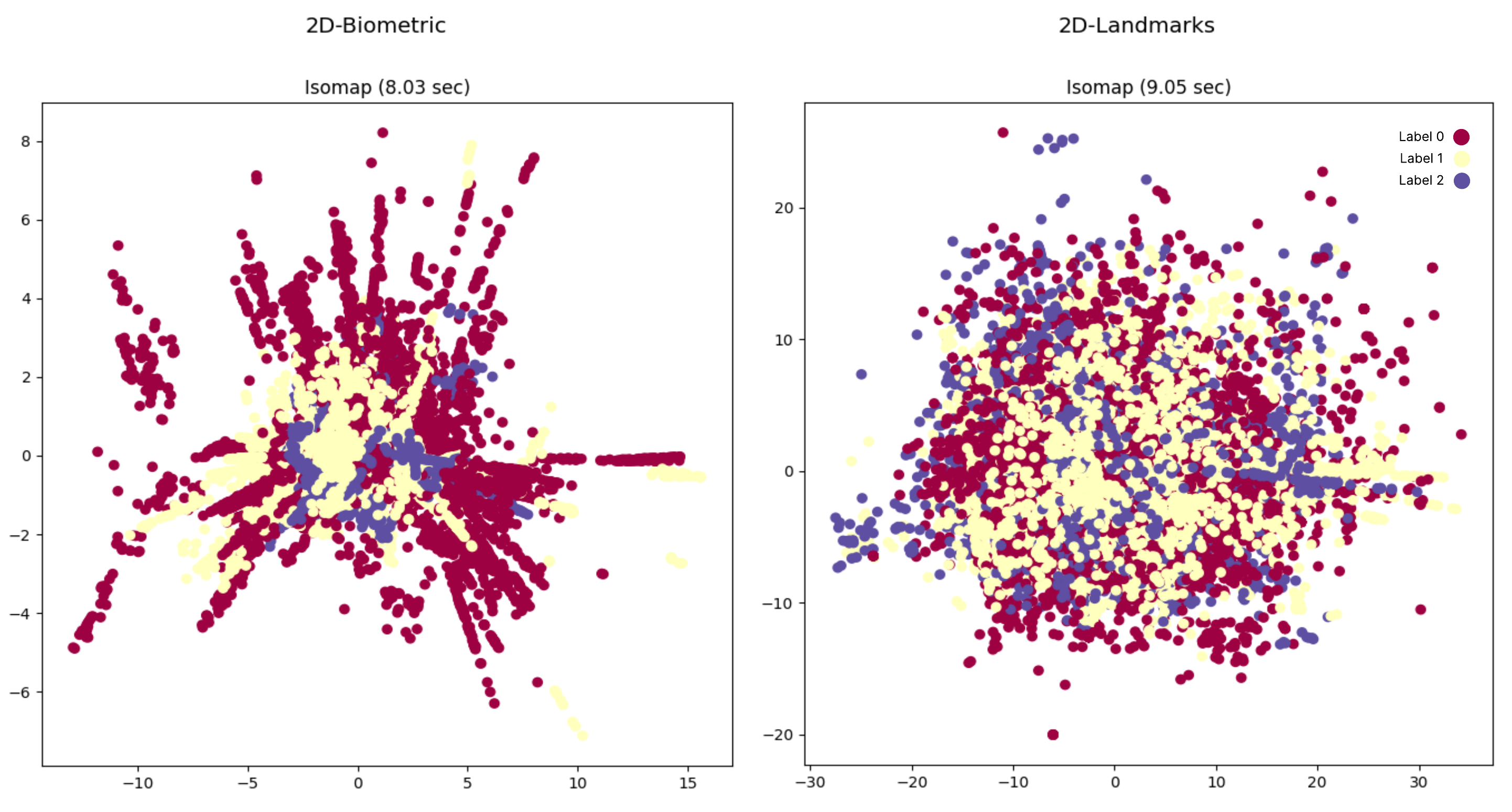}
    \caption{Biometric and Landmarks data visualization using ISO}
    \label{fig:fig7}
\end{figure}

ISO is an extension of MDS. It uses graph distance, which approximates the geodesic distance between all pairs of points \cite{tenenbaum2000global}; in Figure \ref{fig:fig7}, you can see the 2D visualizations of our data obtained using ISO. We used the following parameters for visualization: Number of Neighbors = 10, Number of Components = 2, and random state = 42. The plots obtained through ISO exhibit effective clustering, particularly for biometric signals data, while retaining the global structure of the data. In this study, the accuracy yielded by ISO and MDS was almost the same in an unimodal network for biometric signals.

\subsection{t-Distributed Stochastic Neighbor Embedding (t-SNE)}
\begin{figure}[h!]
    \centering
    \includegraphics[width = \linewidth]{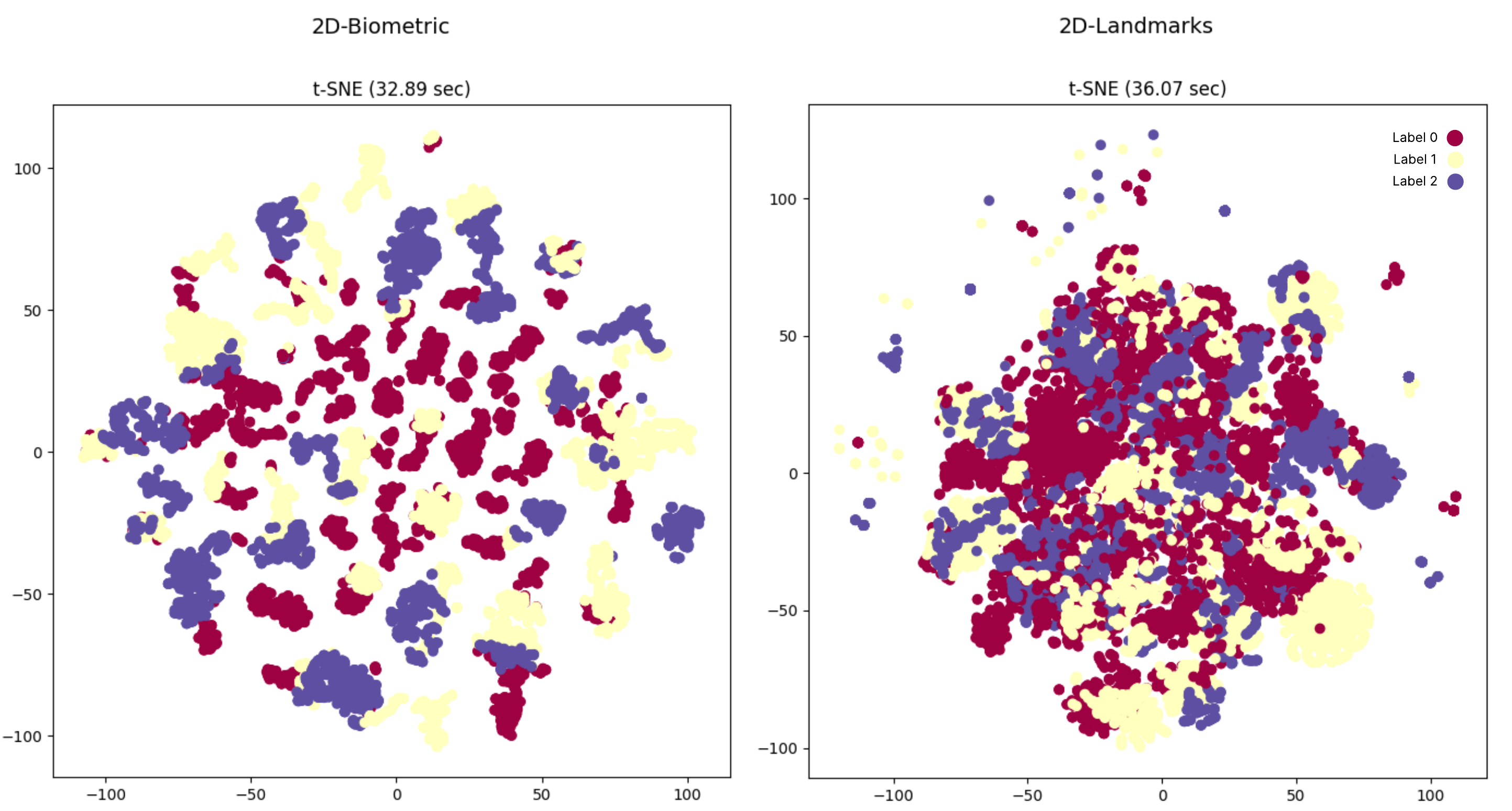}
    \caption{Biometric and Landmarks data visualization using t-SNE}
    \label{fig:fig8}
\end{figure}

t-SNE models each high-dimensional object by a point so that nearby points model similar objects, and dissimilar objects are modeled by distant points with high probability, as described in \cite{van2008visualizing}. We used t-SNE to visualize our data in 2D, as shown in Figure \ref{fig:fig8}. The visualization was performed using the following parameters: perplexity = 30, Number of Components = 2, and random state = 42. The t-SNE plots exhibit promising results in clustering the data, primarily landmarks data, as they capture the local structure of the data. t-SNE ranks as the second method in our study with a high computational cost, and it also corresponds to the second-worst performer in terms of overall performance in multimodal networks. This trade-off between clustering efficacy and computational expense should be carefully considered when choosing t-SNE for dimensionality reduction in similar studies.

\subsection{Principal Component Analysis (PCA)}
\begin{figure}[h!]
    \centering
    \includegraphics[width = \linewidth]{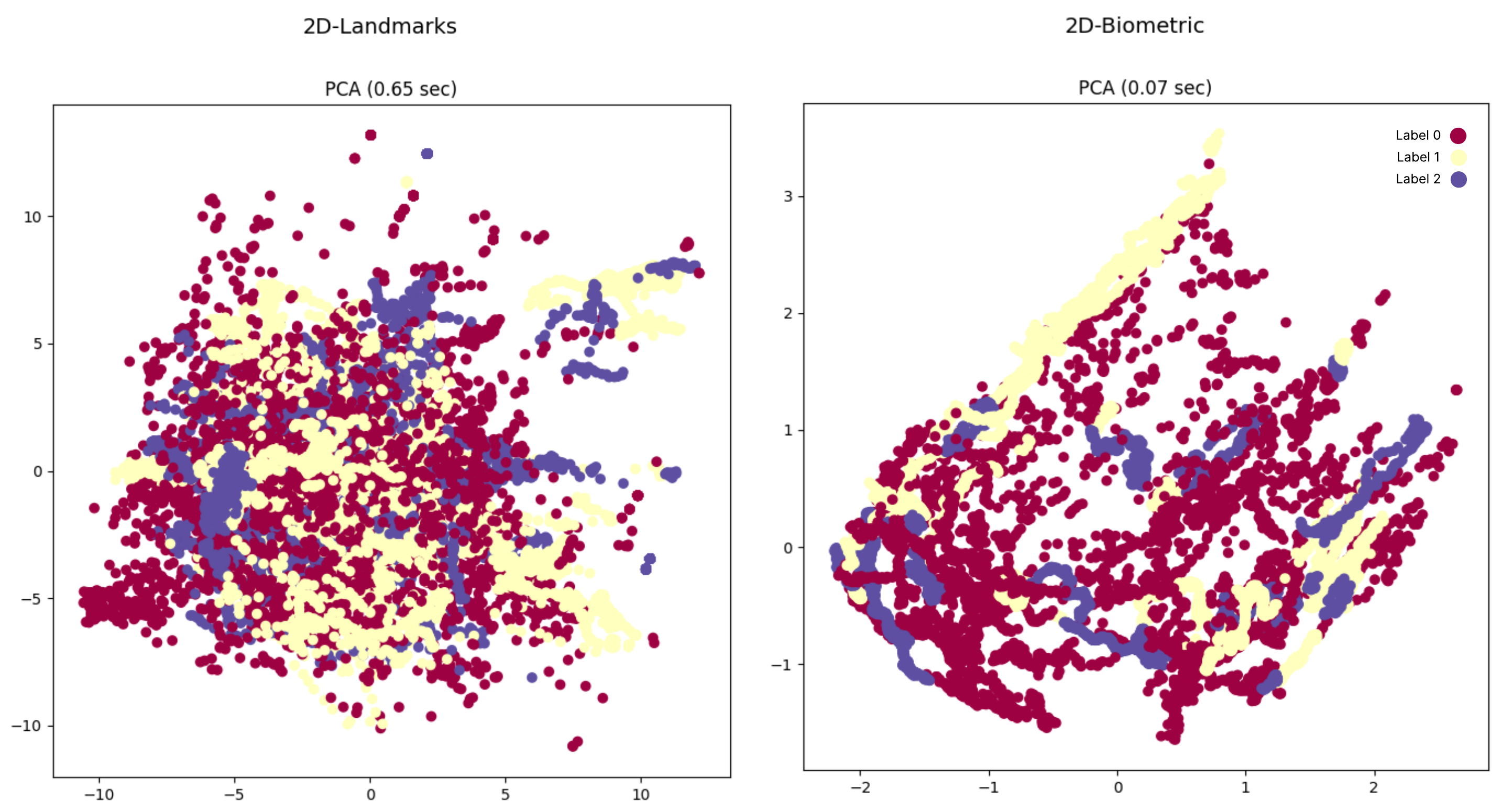}
    \caption{Biometric and Landmarks data visualization using PCA}
    \label{fig:fig10}
\end{figure}

PCA is a linear, well-known method for reducing the dimensions of high-dimensional data. It works by finding the directions of maximum variance and projecting them onto a new subspace with fewer dimensions \cite{jolliffe2016principal}. The figures in \ref{fig:fig10} show how our data is visualized in 2D using PCA. We used the following parameters to generate these visualizations: Number of Components = 2 and random state = 42. PCA plots are effective in clustering, primarily for biometric data and preserving the global structure of the data. The PCA algorithm is more computationally efficient than the manifold methods used in this study. PCA performs the best when using unimodal networks but yields lower performance when it comes to multimodal learning.

\section{Results and Analysis}
We utilized the EmpathicSchool dataset, as introduced in \cite{hosseini2022empathicschool}. This dataset has biometric signals, including heart rate, electrodermal activity, skin temperature, and accelerometer data, alongside video recordings, all of which were employed for stress detection purposes. To efficiently process emotional information without incurring excessive computational costs, we employed the dlib library \cite{king2009dlib} to extract facial landmarks from video data.

For consistency with prior work \cite{hosseini2023multimodal}, we set the signal frequency to 1 Hz. To extract relevant features in both the time and frequency domains, we applied a rolling window of 20 seconds with a 10-second overlap. Within each window, we calculated the average stress level of the user, which served as the stress label. To simplify the stress levels, originally ranging from 0 to 19, we reclassified them into three categories: no-stress (0) for values less than or equal to 6.5, medium-stress (1) for values between 6.5 and 13, and stressful (2) for values exceeding 13.

We applied random down-sampling to address potential data imbalances and minimize bias in our performance evaluation. Furthermore, we independently normalized the minimum and maximum values of each signal for each user. This normalization process helped maintain the original data distribution and mitigated the impact of inter-subject variability.

The resulting dataset featured 1904 features derived from facial landmarks and 175 features from biometric signals. Subsequently, we reduced the data dimension to 49 for facial landmarks and 20 for biometric signals.

We first compared the effect of six manifold learning methods on unimodal networks, and then we applied the manifold learning to the multimodal models.

\subsection{Unimodal Networks}

\begin{figure}[h!]
    \centering
    \includegraphics[width = \linewidth]{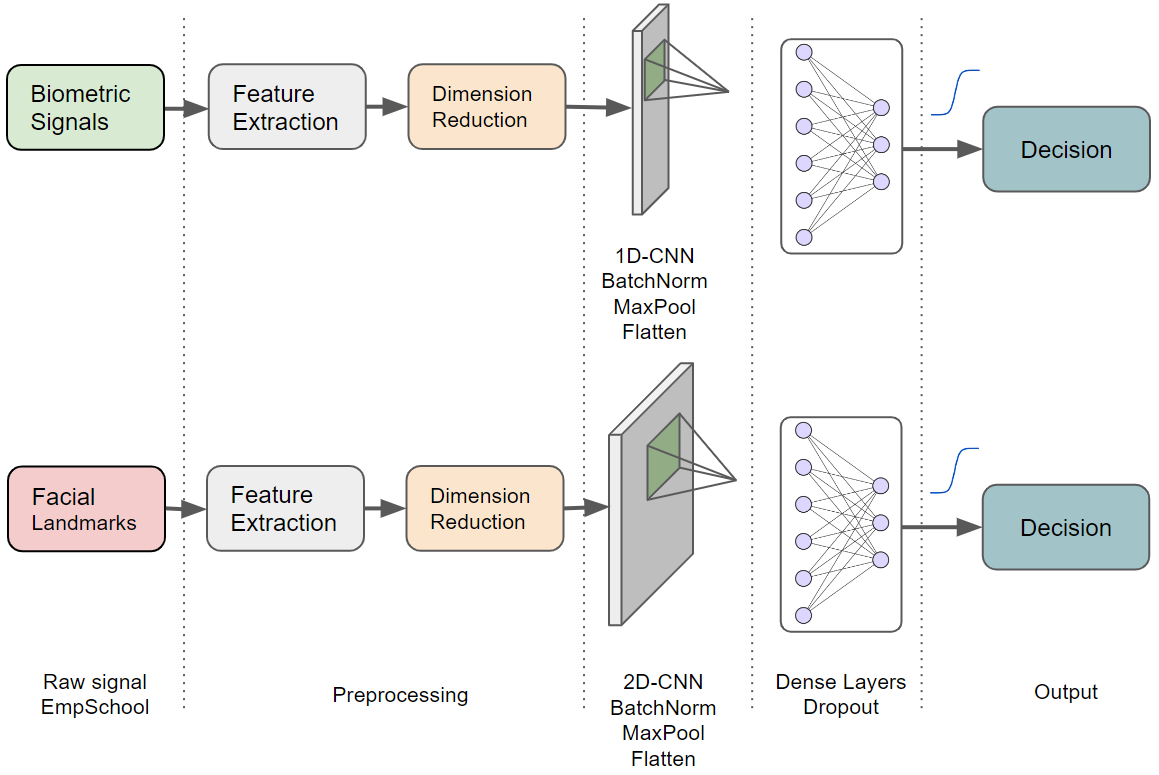}
    \caption{Separate unimodal networks for stress detection using biometric signals and facial landmarks.}
    \label{fig:uni}
\end{figure}

Figure \ref{fig:uni} shows the unimodal network visualization. In these networks, after preprocessing the signals, the data is fed into a $(3\times1)$ CNN for the biometric signals or a $(3\times3)$ CNN for facial landmarks, followed by a $(2\times1)$ or a $(2\times2)$ max-pooling, flattened layer, and DNN layers. The DNN layers consist of a fully connected layer with 16 neurons and a ReLU activation function, followed by a 0.2 dropout. The next layer consists of eight neurons, ReLU activation, L2 regularization of (0.01), followed by 0.2 dropout. The final output layer includes three neurons and a softmax activation function to be consistent with the proposed network. We compare unimodal and multimodal networks to evaluate the effectiveness of multimodal networks in capturing complementary information from different modalities. We investigate the performance computational cost trade-offs of manifold learning using unimodal models.

Table \ref{tab:unibio} and \ref{tab:uniface} show the performance of unimodal deep neural networks using manifold learning. Using LOSO-CV, we trained and tested the deep neural network models for biometric signals (Bio) and facial landmarks (Land) separately. However, manifold learning could not outperform PCA in unimodal networks. Moreover, the PCA has shown lower computational cost than the rest of the dimensionality reduction methods.

\begin{table}[h!]
\centering
\caption{Performance comparison of manifold learning methods on unimodal network using Biometric Signals}
\begin{tabular}{|l|l|l|l|l|l|l|} \hline
method  & LLE   & SE    & MDS   & ISO   & t-SNE & PCA   \\ \hline
Acc(\%) & 75.27 & 64.35 & 81.62 & 80.04 & 48.84 & \textbf{82.22} \\ \hline
Pre(\%) & 70.66 & 66.57 & 76.19 & 74.02 & 46.25 & \textbf{84.13} \\ \hline
Rec(\%) & 73.34 & 59.62 & 80.59 & 74.27 & 46.83 & \textbf{82.10} \\ \hline
F1(\%)  & 69.77 & 54.40 & 76.52 & 72.85 & 42.66 & \textbf{80.11} \\ \hline
\end{tabular}
\label{tab:unibio}
\end{table}

\begin{table}[h!]
\centering
\caption{Performance comparison of manifold learning methods on unimodal network using Facial Landmarks}
\begin{tabular}{|l|l|l|l|l|l|l|} \hline
Method  & LLE   & SE    & MDS   & ISO   & t-SNE & PCA   \\ \hline
Acc(\%) & 59.77 & 36.68 & 69.25 & 55.26 & 61.96 & \textbf{74.13} \\ \hline
Pre(\%) & 58.61 & 44.75 & 64.88 & 53.74 & 67.98 & \textbf{70.56} \\ \hline
Rec(\%) & 59.55 & 45.61 & 65.17 & 52.98 & 57.50 & \textbf{69.63} \\ \hline
F1(\%)  & 54.46 & 31.64 & 62.40 & 46.68 & 54.53 & \textbf{67.52} \\ \hline
\end{tabular}
\label{tab:uniface}
\end{table}

\subsection{Multimodal Early Level Fusion Network}

\begin{figure}[h!]
    \centering
    \includegraphics[width = \linewidth]{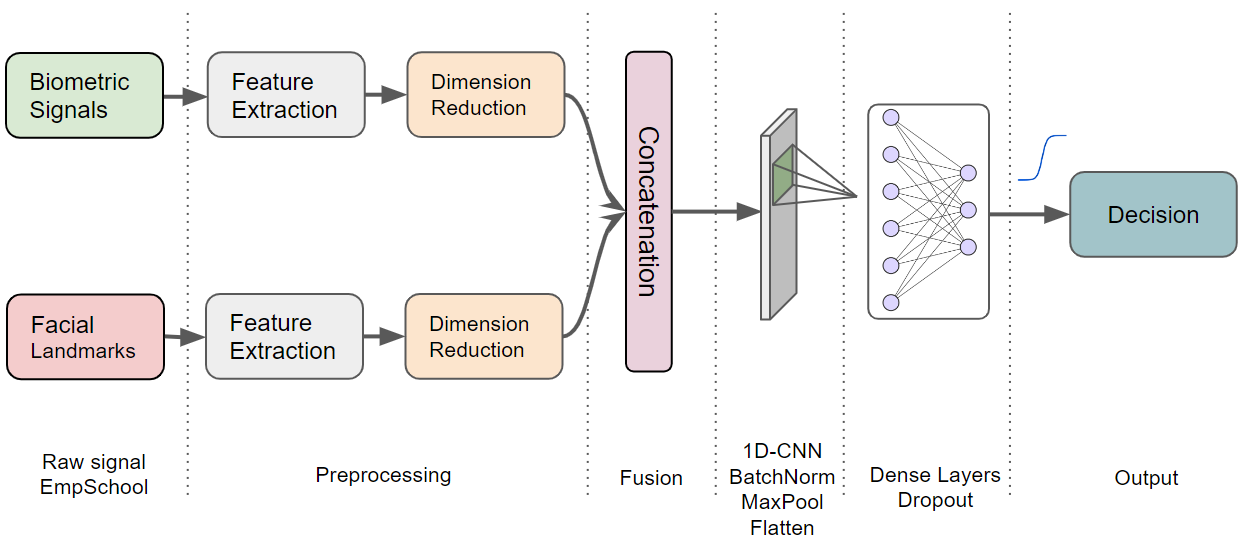}
    \caption{Multimodal early fusion network for stress detection using biometric signals and facial landmarks.}
    \label{fig:early}
\end{figure}

Figure \ref{fig:early} shows the multimodal early-fusion learning network. The dimension reduction method was applied separately to each data type and then concatenated and fed to a 1D-CNN layer followed by max-pooling, flattening, and DNN hidden layers. The DNN layers were consistent throughout all models to facilitate the comparisons. This allows the researchers to compare the effect of adding the manifold learning models on the performance and computational cost of different multimodal learning models. 

Table \ref{tab:early} summarizes the performance of the manifold dimension reduction methods on the early fusion network. LLE showed promising results and outperformed other methods with an accuracy of 90.99\%, followed by ISO and PCA. Manifold learning has enhanced feature integration, while multimodal learning combines features from multiple modalities, leading to a richer data representation \cite{dornaika2020multi}.
LLE performed better than PCA in the early fusion network, attributed to their ability to effectively capture and preserve the intrinsic geometry \cite{jiang2018robust} and structure of the combined biometric signals and facial landmarks.

\begin{table}[h!]
\centering
\caption{Performance comparison of manifold learning methods on Early Fusion network}
\begin{tabular}{|l|l|l|l|l|l|l|} \hline
Method  & LLE   & SE    & MDS   & ISO   & t-SNE & PCA   \\ \hline
Acc(\%)  & \textbf{90.99} & 71.45       & 87.80         & 89.68        & 74.14          & 89.01        \\ \hline
Pre(\%) &86.22 & 75.86       & 83.00        & 87.49        & 68.26          & \textbf{90.84}        \\ \hline
Rec(\%) &86.87 & 64.95       & 85.71        & \textbf{88.67}        & 68.53          & 87.92      \\ \hline
F1(\%)  &86.13 & 61.62       & 82.84        & \textbf{87.29}        & 66.30           & 86.76        \\ \hline
\end{tabular}
\label{tab:early}
\end{table}

\subsection{Multimodal Intermediate Level Fusion Network}
Table \ref{tab:inter} shows the performance of the intermediate-level fusion network using various manifold learning models discussed in Section 4. The results indicate that utilizing the MDS method outperformed other techniques, achieving an accuracy of 96\%. PCA and ISO are followed closely, with accuracies of 93.14\%  and 88.48\%, respectively.
The results show that the MDS accurately represents the pairwise distance in our data and can preserve similarities between data points. Biometric signals and facial landmarks are complex data types that have non-linear relationships between their variables. MDS's ability to capture these relationships becomes crucial in this matter.  Our study showed that choosing a suitable dimension reduction method and the intermediate fusion of convolution layers can result in accurate stress detection compared to early multimodal fusion and unimodal networks. The strength of the 2D-CNN model lies in its ability to extract spatial information from facial landmarks, while the 1D-CNN model excels in capturing temporal dependencies from biometric signals. This integration of spatiotemporal features is crucial for stress detection, as stress manifests in both spatial and temporal aspects. The findings underscore the significance of manifold learning methods in enhancing stress detection performance and cost within the context of an intermediate fusion network.

\begin{table}[h!]
\centering
\caption{Performance comparison of manifold learning methods on Intermediate Fusion network}
\begin{tabular}{|c|c|c|c|c|c|c|}
\hline
Method  & LLE   & SE    & MDS            & ISO   & t-SNE & PCA   \\ \hline
Acc(\%) & 83.16 & 46.44 & \textbf{96.00} & 88.48 & 78.51 & 93.14 \\ \hline
Pre(\%) & 81.74 & 38.56 & \textbf{93.97} & 84.96 & 73.68 & 92.64 \\ \hline
Rec(\%) & 81.24 & 38.47 & \textbf{96.67} & 82.98 & 74.63 & 90.59 \\ \hline
F1(\%)  & 79.04 & 31.57 & \textbf{94.99} & 82.96 & 72.20 & 89.63 \\ \hline
\end{tabular}
\label{tab:inter}
\end{table}

\section{Computational Cost}
Table \ref{tab:com} shows the computational cost of each network in terms of the dimension reduction (DR), training, and prediction times. We used a two-28-core (56 CPUs) node with an Intel(R) Xeon(R) CPU E5-2660 v4 @ 2.00GHz, with L1d, L1i, L2, and L3 cache sizes of 32K, 32K, 256K, and 35840K, respectively. The node was equipped with a P100 Tesla GPU for deep-learning computations.

Manifold dimension reduction methods had higher computational costs than PCA, completing the process in 270 milliseconds for biometric signals (Bio) and 3.28 seconds for facial landmarks (Land). However, the MDS method took the most dimension reduction time, requiring 368 and 483 seconds to reduce the Bio and Land data dimensions, respectively (approximately 100 to 1000 times slower).

\cite{hosseini2023multimodal} explored six different feature selection methods, determining that Lasso regularization outperformed others and yielded the best performance for multimodal deep learning models. The Lasso features selection method took 138 and 1001 seconds to select 20 features from biometric signals and 49 features from facial landmarks, respectively. In contrast, the MDS approach achieved the same dimensional reduction, resulting in higher overall performance in the multimodal methods in a shorter time, showing a 25\% improvement in preprocessing time. This shows the effectiveness of manifold learning in computational cost reduction.

\begin{table*}[h!]
\centering
\caption{Comparing Computational Cost of the Methods in Unimodal, Early, and Intermediate Networks}
\resizebox{\textwidth}{!}{
\begin{tabular}{|c|cc|cccc|cccc|cccc|}
\hline
\multirow{3}{*}{Method} & \multicolumn{2}{c|}{\begin{tabular}[c]{@{}c@{}}DR Time (s)\end{tabular}} & \multicolumn{4}{c|}{Number of Parameters}                                                                                                                               & \multicolumn{4}{c|}{Train  Time (s)}                                                                                               & \multicolumn{4}{c|}{Test  Time (s)}                                                                                                   \\ \cline{2-15} 
 & \multicolumn{1}{c|}{\multirow{2}{*}{Bio}}               & \multirow{2}{*}{Land}               & \multicolumn{2}{c|}{Unimodal}                                                           & \multicolumn{1}{c|}{\multirow{2}{*}{Early}} & \multirow{2}{*}{Intermediate} & \multicolumn{2}{c|}{Unimodal}                      & \multicolumn{1}{c|}{\multirow{2}{*}{Early}} & \multirow{2}{*}{Intermediate} & \multicolumn{2}{c|}{Unimodal}                         & \multicolumn{1}{c|}{\multirow{2}{*}{Early}} & \multirow{2}{*}{Intermediate} \\ \cline{4-5} \cline{8-9} \cline{12-13}
 & \multicolumn{1}{c|}{}                                   &                                     & \multicolumn{1}{c|}{Bio}                    & \multicolumn{1}{c|}{Land}                   & \multicolumn{1}{c|}{}                       &                               & \multicolumn{1}{c|}{Bio} & \multicolumn{1}{c|}{Land} & \multicolumn{1}{c|}{}                       &                               & \multicolumn{1}{c|}{Bio}   & \multicolumn{1}{c|}{Land}  & \multicolumn{1}{c|}{}                       &                               \\ \hline
LLE                     & \multicolumn{1}{c|}{7.99}                               & 10.10                               & \multicolumn{1}{c|}{\multirow{6}{*}{5,043}} & \multicolumn{1}{c|}{\multirow{6}{*}{2,675}} & \multicolumn{1}{c|}{\multirow{6}{*}{5,043}} & \multirow{6}{*}{82,099}       & \multicolumn{1}{c|}{413} & \multicolumn{1}{c|}{346}  & \multicolumn{1}{c|}{402}                    & 674                           & \multicolumn{1}{c|}{0.084} & \multicolumn{1}{c|}{0.080} & \multicolumn{1}{c|}{0.081}                  & 0.394                         \\ \cline{1-3} \cline{8-15} 
SE                      & \multicolumn{1}{c|}{118}                                & 42.70                               & \multicolumn{1}{c|}{}                       & \multicolumn{1}{c|}{}                       & \multicolumn{1}{c|}{}                       &                               & \multicolumn{1}{c|}{258} & \multicolumn{1}{c|}{236}  & \multicolumn{1}{c|}{343}                    & 391                           & \multicolumn{1}{c|}{0.079} & \multicolumn{1}{c|}{0.064} & \multicolumn{1}{c|}{0.082}                  & 0.387                         \\ \cline{1-3} \cline{8-15} 
MDS                     & \multicolumn{1}{c|}{368}                                & 483                                 & \multicolumn{1}{c|}{}                       & \multicolumn{1}{c|}{}                       & \multicolumn{1}{c|}{}                       &                               & \multicolumn{1}{c|}{350} & \multicolumn{1}{c|}{254}  & \multicolumn{1}{c|}{413}                    & 466                           & \multicolumn{1}{c|}{0.077} & \multicolumn{1}{c|}{0.108} & \multicolumn{1}{c|}{0.097}                  & 0.403                         \\ \cline{1-3} \cline{8-15} 
ISO                     & \multicolumn{1}{c|}{8.44}                               & 12.60                               & \multicolumn{1}{c|}{}                       & \multicolumn{1}{c|}{}                       & \multicolumn{1}{c|}{}                       &                               & \multicolumn{1}{c|}{359} & \multicolumn{1}{c|}{281}  & \multicolumn{1}{c|}{389}                    & 510                           & \multicolumn{1}{c|}{0.078} & \multicolumn{1}{c|}{0.065} & \multicolumn{1}{c|}{0.084}                  & 0.389                         \\ \cline{1-3} \cline{8-15} 
t-SNE                   & \multicolumn{1}{c|}{62}                                 & 414                                 & \multicolumn{1}{c|}{}                       & \multicolumn{1}{c|}{}                       & \multicolumn{1}{c|}{}                       &                               & \multicolumn{1}{c|}{298} & \multicolumn{1}{c|}{282}  & \multicolumn{1}{c|}{435}                    & 592                           & \multicolumn{1}{c|}{0.078} & \multicolumn{1}{c|}{0.065} & \multicolumn{1}{c|}{0.080}                  & 0.375                         \\ \cline{1-3} \cline{8-15} 
PCA                     & \multicolumn{1}{c|}{0.27}                               & 3.28                                & \multicolumn{1}{c|}{}                       & \multicolumn{1}{c|}{}                       & \multicolumn{1}{c|}{}                       &                               & \multicolumn{1}{c|}{343} & \multicolumn{1}{c|}{238}  & \multicolumn{1}{c|}{326}                    & 468                           & \multicolumn{1}{c|}{0.077} & \multicolumn{1}{c|}{0.064} & \multicolumn{1}{c|}{0.122}                  & 0.383                         \\ \hline
\end{tabular}
}
\label{tab:com}
\end{table*}

\subsection{Effect of adding extra non-linearity using 1D-CNN}
Adding an extra 1D-CNN layer, followed by a max-pooling, right after data fusion yields performance improvement by adding non-linearity to the model, enhancing the average performance of the models by an increase of 7.41\% in accuracy, 6.18\% in precision, 9.23\% in recall, and 9.42\% in f1-score. This final 1D-CNN layer plays a significant role in capturing higher-level patterns by combining extracted features from both modalities \cite{radhika2021deep}. 

 \section{Discussion and Future works}
 In this paper, we presented a streamlined intermediate-level fusion network that leverages manifold learning techniques to enhance stress detection. Within this model, manifold-based dimension reduction is separately applied to each signal in the initial stages. This unique approach enables the extraction of meaningful representations from biometric signals and facial landmarks, both of which possess intricate and non-linear relationships, independently. These representations are later combined to inform the stress detection process. It's worth noting that this approach can be customized to different levels for various network configurations.

In this study, we demonstrate the efficacy of non-linear dimension reduction methods within intermediate multimodal neural networks when compared to early fusion and unimodal models. Notably, our model, incorporating Multidimensional Scaling (MDS), outperforms the other models, while Locally Linear Embedding (LLE) strikes a balance between performance and computational costs. Principal Component Analysis (PCA) offers higher processing speed due to its efficiency. However, the incorporation of manifold learning techniques results in a substantial 41.69\% reduction in errors (from 93.14\% to 96\%) compared to using PCA.

We plan to investigate the effect of implementing manifold learning \ the multimodal learning network at various levels of fusion: multimodal early, late, and intermediate-level fusion models. Another research area is investigating how manifold learning preserves complementary information to enhance multimodal learning for stress detection.

\section{Summary and Conclusions}
In summary, this study introduced a lightweight intermediate fusion network that employs manifold learning for dimensionality reduction. We systematically investigated the impact of incorporating manifold learning in the preprocessing phase across unimodal, early, and intermediate-fusion networks. Remarkably, manifold learning yielded a substantial 41.69\% reduction in errors when compared to Principal Component Analysis (PCA) within intermediate-level fusion networks. Our analysis involved the examination of various manifold dimension reduction methods, assessing their effects on stress detection performance and computational costs. We evaluated six manifold learning models, including LLE, SE, MDS, ISO, t-SNE, and PCA.

The results underscored the superiority of the intermediate-level fusion model that integrated the MDS manifold learning dimensionality reduction method, achieving an impressive 96\% accuracy in the Leave-One-Subject-Out Cross-Validation (LOSO-CV) paradigm. Furthermore, the implementation of data balancing through down-sampling played a pivotal role in reducing dimension reduction and train-test times, ensuring that performance assessments remained unbiased in the context of stress detection.

Additionally, the inclusion of an extra 1D-CNN layer had a substantial positive impact on the model's performance, leading to a 7.41\% increase in accuracy, a 6.18\% boost in precision, a 9.23\% improvement in recall, and a 9.42\% enhancement in the f1-score. These findings collectively highlight the significance of manifold learning, data balancing, and architectural enhancements in advancing the state-of-the-art in stress detection techniques.

\printbibliography

@article{dornaika2020multi,
  title={Multi-layer manifold learning with feature selection},
  author={Dornaika, Fadi},
  journal={Applied Intelligence},
  volume={50},
  number={6},
  pages={1859--1871},
  year={2020},
  publisher={Springer}
}

@article{walambe2023employing,
  title={Employing Multimodal Machine Learning for Stress Detection},
  author={Walambe, Rahee and Nayak, Pranav and Bhardwaj, Ashmit and Kotecha, Ketan},
  journal={arXiv preprint arXiv:2306.09385},
  year={2023}
}

@article{zhu2019multimodal, title={Multimodal brain network jointly construction and fusion for diagnosis of epilepsy}, author={Zhu, Qi and Yang, Jing and Xu, Bingliang and Hou, Zhenghua and Sun, Liang and Zhang, Daoqiang}, journal={Frontiers in Neuroscience}, volume={15}, pages={734711}, year={2021}, publisher={Frontiers} }

@article{song2020multi, title={Multi-modal feature selection with self-expression topological manifold learning}, author={Song, Chaofan and Liu, Tongqiang and Wang, Huan and Shi, Haifeng and Jiao, Zhuqing}, journal={Mathematical Biosciences and Engineering}, volume={18}, number={5}, pages={6468–6482}, year={2021}, publisher={American Institute of Mathematical Sciences} }

@article{nguyen2019ten, title={Ten quick tips for effective dimensionality reduction}, author={Nguyen, Lam Ho and Holmes, Susan}, journal={PLOS Computational Biology}, volume={15}, number={6}, pages={e1006907}, year={2019}, publisher={Public Library of Science} }

@article{al2019redundancy, title={Redundancy detection and removal in big data: A comprehensive review}, author={Al-Saadi, Saad and Al-Dujaili, Abdullah and Li, Jingpeng}, journal={IEEE Access}, volume={7}, pages={106328–106344}, year={2019}, publisher={IEEE} }

@article{dutt2022shared, title={Shared Manifold Learning Using a Triplet Network for Multiple Sensor Translation and Fusion With Missing Data}, author={Dutt, Aditya and Zare, Alina and Gader, Paul}, journal={IEEE Journal of Selected Topics in Applied Earth Observations and Remote Sensing}, volume={15}, pages={9439–9456}, year={2022}, publisher={IEEE} }

@article{zhu2023survey, title={A Survey on Model Compression for Large Language Models}, author={Zhu, Xunyu and Li, Jian and Liu, Yong and Ma, Can and Wang, Weiping}, journal={arXiv preprint arXiv:2308.07633}, year={2023} }

@article{al2021review, title={Review of deep learning: concepts, CNN architectures, challenges, applications, future directions}, author={Al-Dujaili, Abdullah and Al-Saadi, Saad}, journal={Journal of Big Data}, volume={8}, number={1}, pages={1–49}, year={2021}, publisher={Springer} }

@inproceedings{kapsouras2020deep, title={A Deep Learning Approach Towards Multimodal Stress Detection}, author={Kapsouras, Ioannis and Nikolopoulos, Spiros and Kompatsiaris, Ioannis}, booktitle={Proceedings of the 2nd International Workshop on Affective Computing}, pages={1–8}, year={2020} }

@article{zhao2023stress, title={Stress Detection via Multimodal Multitemporal-Scale Fusion: A Hybrid of Handcrafted and Deep Learning Features}, author={Zhao, Shengdong and Li, Xiang and Wang, Zhi and Wang, Yiqiang and Chen, Feng and Chen, Huiling and Zhang, Zhenyu}, journal={IEEE Transactions on Affective Computing}, year={2023}, publisher={IEEE} }

@inproceedings{zhang2019multimodal, title={Multimodal Stress Detection from Multiple Heterogeneous Data Sources}, author={Zhang, Zhenyu and Wang, Zhi and Zhao, Shengdong and Li, Xiang and Wang, Yiqiang and Chen, Feng and Chen, Huiling}, booktitle={Proceedings of the 27th ACM International Conference on Multimedia}, pages={2410–2418}, year={2019} }

@article{li2020stress,
  title={Stress detection using deep neural networks},
  author={Li, Russell and Liu, Zhandong},
  journal={BMC Medical Informatics and Decision Making},
  volume={20},
  pages={1--10},
  year={2020},
  publisher={Springer}
}

@article{hosseini2022multimodal,
  title={A multimodal sensor dataset for continuous stress detection of nurses in a hospital},
  author={Hosseini, Seyedmajid and Gottumukkala, Raju and Katragadda, Satya and Bhupatiraju, Ravi Teja and Ashkar, Ziad and Borst, Christoph W and Cochran, Kenneth},
  journal={Scientific Data},
  volume={9},
  number={1},
  pages={255},
  year={2022},
  publisher={Nature Publishing Group UK London}
}

@article{zhang2022multimodal,
  title={Multimodal emotion recognition based on manifold learning and convolution neural network},
  author={Zhang, Yong and Cheng, Cheng and Zhang, YiDie},
  journal={Multimedia Tools and Applications},
  volume={81},
  number={23},
  pages={33253--33268},
  year={2022},
  publisher={Springer}
}

@article{king2009dlib,
  title={Dlib-ml: A Machine Learning Toolkit},
  author={King, Davis E},
  journal={Journal of Machine Learning Research},
  volume={10},
  pages={1755--1758},
  year={2009}
}

@article{hosseini2022empathicschool,
  title={EmpathicSchool: A multimodal dataset for real-time facial expressions and physiological data analysis under different stress conditions},
  author={Hosseini, Majid and Sohrab, Fahad and Gottumukkala, Raju and Bhupatiraju, Ravi Teja and Katragadda, Satya and Raitoharju, Jenni and Iosifidis, Alexandros and Gabbouj, Moncef},
  journal={arXiv preprint arXiv:2209.13542},
  year={2022}
}

@article{roweis2000nonlinear,
  title={Nonlinear dimensionality reduction by locally linear embedding},
  author={Roweis, Sam T and Saul, Lawrence K},
  journal={science},
  volume={290},
  number={5500},
  pages={2323--2326},
  year={2000},
  publisher={American Association for the Advancement of Science}
}

@article{belkin2003laplacian,
  title={Laplacian eigenmaps for dimensionality reduction and data representation},
  author={Belkin, Mikhail and Niyogi, Partha},
  journal={Neural computation},
  volume={15},
  number={6},
  pages={1373--1396},
  year={2003},
  publisher={MIT Press}
}

@book{borg2005modern,
  title={Modern multidimensional scaling: Theory and applications},
  author={Borg, Ingwer and Groenen, Patrick JF},
  year={2005},
  publisher={Springer Science \& Business Media}
}

@article{tenenbaum2000global,
  title={A global geometric framework for nonlinear dimensionality reduction},
  author={Tenenbaum, Joshua B and Silva, Vin de and Langford, John C},
  journal={science},
  volume={290},
  number={5500},
  pages={2319--2323},
  year={2000},
  publisher={American Association for the Advancement of Science}
}

@article{van2008visualizing,
  title={Visualizing data using t-SNE.},
  author={Van der Maaten, Laurens and Hinton, Geoffrey},
  journal={Journal of machine learning research},
  volume={9},
  number={11},
  year={2008}
}

@article{jolliffe2016principal,
  title={Principal component analysis: a review and recent developments},
  author={Jolliffe, Ian T and Cadima, Jorge},
  journal={Philosophical transactions of the royal society A: Mathematical, Physical and Engineering Sciences},
  volume={374},
  number={2065},
  pages={20150202},
  year={2016},
  publisher={The Royal Society Publishing}
}

@article{huang2012nonlinear,
  title={On nonlinear dimensionality reduction for face recognition},
  author={Huang, Weilin and Yin, Hujun},
  journal={Image and Vision Computing},
  volume={30},
  number={4-5},
  pages={355--366},
  year={2012},
  publisher={Elsevier}
}

@inproceedings{akhloufi2009multispectral,
  title={Multispectral face recognition using non linear dimensionality reduction},
  author={Akhloufi, Moulay A and Bendada, Abdelhakim and Batsale, Jean-Christophe},
  booktitle={Visual Information Processing XVIII},
  volume={7341},
  pages={152--161},
  year={2009},
  organization={SPIE}
}

@article{hosseini2023multimodal,
  title={Multimodal Stress Detection Using Facial Landmarks and Biometric Signals},
  author={Hosseini, Majid and Bodaghi, Morteza and Bhupatiraju, Ravi Teja and Maida, Anthony and Gottumukkala, Raju},
  journal={arXiv preprint arXiv:2311.03606},
  year={2023}
}

@article{hoffmann2020learning,
  title={Learning multi-modal image registration without real data},
  author={Hoffmann, Malte and Billot, Benjamin and Iglesias, Juan Eugenio and Fischl, Bruce and Dalca, Adrian V},
  journal={arXiv preprint arXiv:2004.10282},
  year={2020}
}

@article{ugalde2015computational,
  title={Computational cost improvement of neural network models in black box nonlinear system identification},
  author={Ugalde, Hector M Romero and Carmona, Jean-Claude and Reyes-Reyes, Juan and Alvarado, Victor M and Mantilla, Juan},
  journal={Neurocomputing},
  volume={166},
  pages={96--108},
  year={2015},
  publisher={Elsevier}
}

@article{wang2012folded,
  title={A folded neural network autoencoder for dimensionality reduction},
  author={Wang, Jing and He, Haibo and Prokhorov, Danil V},
  journal={Procedia Computer Science},
  volume={13},
  pages={120--127},
  year={2012},
  publisher={Elsevier}
}

@article{murugan2017regularization,
  title={Regularization and optimization strategies in deep convolutional neural network},
  author={Murugan, Pushparaja and Durairaj, Shanmugasundaram},
  journal={arXiv preprint arXiv:1712.04711},
  year={2017}
}

@inproceedings{radhika2021deep,
  title={Deep multimodal fusion for subject-independent stress detection},
  author={Radhika, K and Oruganti, V Ramana Murthy},
  booktitle={2021 11th International Conference on Cloud Computing, Data Science \& Engineering (Confluence)},
  pages={105--109},
  year={2021},
  organization={IEEE}
}

@inproceedings{wu2022fusion,
  title={Fusion of Physiological and Behavioural Signals on SPD Manifolds with Application to Stress and Pain Detection},
  author={Wu, Yujin and Daoudi, Mohamed and Amad, Ali and Sparrow, Laurent and D’Hondt, Fabien},
  booktitle={2022 IEEE International Conference on Systems, Man, and Cybernetics (SMC)},
  pages={2949--2955},
  year={2022},
  organization={IEEE}
}

@article{seo2022deep,
  title={Deep learning approach for detecting work-related stress using multimodal signals},
  author={Seo, Wonju and Kim, Namho and Park, Cheolsoo and Park, Sung-Min},
  journal={IEEE Sensors Journal},
  volume={22},
  number={12},
  pages={11892--11902},
  year={2022},
  publisher={IEEE}
}

@article{zhang2022real,
  title={Real-time mental stress detection using multimodality expressions with a deep learning framework},
  author={Zhang, Jing and Yin, Hang and Zhang, Jiayu and Yang, Gang and Qin, Jing and He, Ling},
  journal={Frontiers in Neuroscience},
  volume={16},
  pages={947168},
  year={2022},
  publisher={Frontiers}
}

@article{zali2023deep,
  title={Deep time-frequency features and semi-supervised dimension reduction for subject-independent emotion recognition from multi-channel EEG signals},
  author={Zali-Vargahan, Behrooz and Charmin, Asghar and Kalbkhani, Hashem and Barghandan, Saeed},
  journal={Biomedical Signal Processing and Control},
  volume={85},
  pages={104806},
  year={2023},
  publisher={Elsevier}
}

@book{ma2011manifold,
  title={Manifold learning theory and applications},
  author={Ma, Yunqian and Fu, Yun},
  year={2011},
  publisher={CRC press}
}

@article{she2023cross,
  title={Cross-subject EEG emotion recognition using multi-source domain manifold feature selection},
  author={She, Qingshan and Shi, Xinsheng and Fang, Feng and Ma, Yuliang and Zhang, Yingchun},
  journal={Computers in Biology and Medicine},
  volume={159},
  pages={106860},
  year={2023},
  publisher={Elsevier}
}

@article{sreevidya2022elder,
  title={Elder emotion classification through multimodal fusion of intermediate layers and cross-modal transfer learning},
  author={Sreevidya, P and Veni, S and Ramana Murthy, OV},
  journal={Signal, image and video processing},
  volume={16},
  number={5},
  pages={1281--1288},
  year={2022},
  publisher={Springer}
}

@article{jiang2018robust,
  title={Robust data representation using locally linear embedding guided PCA},
  author={Jiang, Bo and Ding, Chris and Luo, Bin},
  journal={Neurocomputing},
  volume={275},
  pages={523--532},
  year={2018},
  publisher={Elsevier}
}

\end{document}